# Nsanku: Evaluating Zero-Shot Translation Performance of LLMs for Ghanaian Languages

Stephen E. Moore[1,2,3], Mich-Seth Owusu[2], Akwasi Asare[2,3], Lawrence Adu Gyamfi[2,3], Paul Azunre[2,3], Joel Budu[2,4], Jonathan Asiamah[2], Elias Dzobo[2], Kelvin Newman[2], Edmund O. Benefo[2], Gerhardt Datsomor[2], Onesimus Addo Appiah[2], Ama Branoa Banful[2], Lucas Woedem Kpatah[2], Saani Mustapha Deishini[2], John Ayernor[2]

[1]Department of Mathematics, University of Cape Coast, Cape Coast, Ghana

[2]Ghana Natural Language Processing, Cape Coast, Cape Coast, Ghana

[3]Khaya AI LTD Ghana

**Abstract**: Large language models (LLMs) have demonstrated impressive multilingual capabilities for well-resourced languages, yet their performance on low-resource African languages remains poorly understood and largely unevaluated. This paper presents Nsanku, a systematic benchmark that evaluates the zero-shot machine translation performance of 19 open-weight and proprietary LLMs across 43 Ghanaian languages paired with English. Evaluation sentences were sourced from the YouVersion Bible platform, providing 300 sentence pairs per language. Two complementary automatic metrics are employed: Bilingual Evaluation Understudy (BLEU) and Character n-gram F-Score (chrF), alongside an average accuracy score and a cross-language consistency dimension. Nsanku represents the most comprehensive LLM translation evaluation for Ghanaian languages conducted to date. Results show that gemini-2.5-flash achieves the highest overall average score of 26.88 (BLEU: 24.60, chrF: 29.16), followed by claude-sonnet-4-5 at 24.87 (BLEU: 22.46, chrF: 27.28) and gpt-4.1 at 23.20 (BLEU: 21.15, chrF: 25.24). Among open-weight models, kimi-k2-instruct-0905 leads at an average score of 20.87. A critical finding from the consistency analysis is that no model and no language reached the Leaders quadrant of high performance and high consistency simultaneously, indicating that current LLMs are not yet reliably usable for Ghanaian language translation at scale. Siwu achieved the highest per-language average score at 25.73 while Nkonya scored lowest at 11.65. Nsanku establishes a publicly available, community-extensible evaluation infrastructure for African language NLP research.

**Corresponding Author:** stephen.moore@ucc.edu.gh

## 1. INTRODUCTION

Ghana is home to over eighty (80) documented languages spanning several major linguistic families, including Kwa, Gur, Grusi, and Mande [1, 2]. These languages are spoken daily by millions of Ghanaians in markets, homes, churches, schools, and courts, yet they remain almost entirely absent from the tools and technologies that define modern natural language processing. The rapid rise of large language models (LLMs) has reshaped what is possible in machine translation: systems such as GPT-4, Gemini, and other open-weight alternatives from Meta, Mistral, and Alibaba can now produce grammatically fluent translations between hundreds of languages without ever being explicitly trained on a parallel corpus for those language pairs [3], [4]. For well-resourced languages such as French, Mandarin, and Spanish, LLMs have become powerful and widely deployed translation tools [4,5,6,7]. For Ghanaian languages, however, it remains entirely unclear whether these same systems are capable, partially capable, or fundamentally inadequate because no systematic evaluation has ever been conducted at scale.

This gap is not merely academic. The inability to access reliable machine translation in a native language has real consequences: citizens may be unable to use digital health services, access government information, or benefit from educational technology that is designed entirely around English [5, 8, 9]. As LLM-powered translation tools are deployed in consumer applications, healthcare systems, and public services across Africa, the communities most in need of equitable language technology are precisely those for whom the technology has never been evaluated. Without a systematic benchmark, developers cannot know which models to deploy, policymakers cannot make evidence-based decisions, and researchers cannot identify where the most critical capability gaps lie. Nsanku was built to fill that void.

Nsanku, meaning musical instruments in the Akan language, is a Ghana NLP initiative that provides the first systematic, large-scale evaluation of LLM machine translation performance across 43 Ghanaian languages. The benchmark evaluates 19 LLMs spanning frontier proprietary systems and a broad range of open-weight alternatives under a zero-shot protocol using parallel sentence pairs sourced from the YouVersion Bible platform [10]. Two established automatic metrics are employed: Bilingual Evaluation Understudy, BLEU, and Character F-Score, or chrF [11, 12] , alongside an average accuracy score and a cross-lingual consistency dimension that reveals whether a model's capability is stable or concentrated on a small subset of languages. By releasing all data, code, and results openly, Nsanku is designed to serve not only as a one-time snapshot but as a living, community-extensible infrastructure that the research community can build on, extend, and update as new models emerge.

The scale of the evaluation is itself a significant contribution. Major multilingual benchmarks such as FLORES-200 [13] and XTREME [14] cover at most one Ghanaian language, Twi, leaving 42 of the 43 languages evaluated here entirely absent from prior systematic NLP evaluation. MasakhaNEWS [3] and related African NLP efforts have made important strides in building datasets for a small number of prominent African languages, but the majority of Ghanaian languages remain without any publicly available benchmark. Nsanku addresses this directly, establishing for the first time a common evaluation floor across the full documented breadth of Ghana's language landscape, from widely spoken languages such as Twi, Ewe, and Dagbani to smaller varieties such as Siwu, Nkonya, and Tumulung Sisaala.

This study is guided by three interconnected aims. First, it evaluates and compares the zero-shot machine translation capabilities of 19 leading LLMs across 43 Ghanaian languages, measuring

**Corresponding Author:** stephen.moore@ucc.edu.gh

performance using BLEU, chrF, and an average accuracy score. Second, it quantifies the performance disparity between frontier proprietary models and open-weight alternatives, examining whether open-weight development has begun to close the gap for these underrepresented languages. Third, it assesses model reliability through a cross-lingual consistency analysis, determining whether translation accuracy is stable across the full language set or disproportionately concentrated on languages with larger digital footprints, such as Twi.

The remainder of this paper is structured as follows. Section 2 situates the work within the existing literature on machine translation evaluation, LLM multilingual capabilities, low-resource African language NLP, and prior benchmarking efforts. Section 3 describes the full methodology, covering the data collection and alignment pipeline, the model evaluation architecture, the 43 evaluated languages, the sampling and inference protocol, and the metric and reporting framework. Section 4 presents and discusses results at both the model level and the language level, including a direct metric comparison and the consistency quadrant analysis. Section 5 acknowledges the key limitations of the study, and Section 6 concludes with a summary of contributions and directions for future work.

## 2. Literature Review

### 2.1 Evaluation Metrics for Machine Translation

BLEU, proposed by Papineni et. al. [15], measures n-gram precision between candidate and reference translations with a brevity penalty and became the de facto standard for MT evaluation due to its speed and reproducibility. Its limitations are equally well documented: sensitivity to tokenisation choices, blindness to synonyms and paraphrases, and poor performance at the segment level. The WMT22 Metrics Shared Task as discussed by [16] evaluated a large collection of metrics against expert human ratings across four domains and found that neural-based learned metrics significantly outperformed BLEU, which ranked near the bottom across all evaluation scenarios. These findings inform the metric choices in Nsanku and underline the importance of reporting BLEU alongside a complementary metric rather than in isolation.

Character n-gram F-Score (chrF) computes the F-score of character n-gram matches between hypothesis and reference, see, e.g. [11]. Operating at the sub-word level, it is more robust to morphological variation and does not require tokenisation decisions, making it particularly appropriate for morphologically rich languages. For Nsanku, both BLEU and chrF are reported together: BLEU enables comparison with prior African language MT work, while chrF is better suited to the morphological complexity of Kwa, Gur, and related Ghanaian language families. Beyond accuracy, evaluation is increasingly understood to require a consistency dimension, since it is insufficient for a model to produce high-quality translations on some occasions without doing so reliably across prompts and linguistic structures [17, 18].

### 2.2 LLM Capabilities for Machine Translation

Hendy et al. [19] conducted a comprehensive evaluation of GPT models across eighteen (18) translation directions and found competitive quality for high-resource languages but limited capability for low-resource ones, with zero-shot settings particularly disadvantaging underrepresented languages. Robinson and colleagues [4] tested ChatGPT on 204 languages using FLORES-200 and found it underperformed traditional MT for 84.1% of evaluated languages, outperforming NLLB on 47% of high-resource languages but only 6% of

**Corresponding Author:** stephen.moore@ucc.edu.gh

low-resource ones, with African languages especially disadvantaged. Zhu et al. [20] confirmed that GPT-4 beats the NLLB baseline in 40.91 percent of translation directions but still trails commercial systems on low-resource languages. Together these studies establish a clear pattern: LLM translation performance tracks resource availability closely, and the zero-shot setting places maximum pressure on whatever language knowledge models have acquired implicitly during pre-training.

Research has further shown that multilingual models often prioritise high-resource languages during training [21]. Alabi et al [22] found that the Twi vocabulary in fastText Wikipedia embeddings contained only 935 words, with the Yoruba vocabulary dominated by words from other languages including English, French, and Arabic, structurally disadvantageous to these languages regardless of model scale. This finding provides direct context for the modest absolute scores observed in Nsanku.

### 2.3 African and Ghanaian Language NLP

Nekoto et al. [23] published a participatory research case study producing MT benchmarks for over thirty African languages, arguing that low-resourcedness reflects systemic resource scarcity in the societies from which speakers come, not merely a technical data problem. For Ghanaian languages specifically, Azunre et al.[2], [24] produced the first substantial parallel corpus for English and Akuapem Twi with 25,421 sentence pairs, and Agyei et al. [25] contributed the multi-domain Twi-2-ENG corpus drawing on parliamentary transcripts, news portals, the Bible, and social media. Adelani et al. [3] demonstrated through MasakhaNEWS that few-shot LLM approaches can achieve more than 90 percent of fully supervised performance for African language classification, though generation tasks such as MT remain substantially harder. Mensah et al. [26] found strong domain dependency in Akan ASR, with models performing optimally only within their training domain, a finding directly relevant to Nsanku's scriptural evaluation corpus. In addition to broader multilingual benchmarks, Ghana-specific resources such as the *Ndwom* and *Ayoo* MIR datasets [27, 28], together with parallel corpora released by GhanaNLP [1], have begun addressing the underrepresentation of Ghanaian languages and cultural content in NLP, machine translation, speech technology, and music information retrieval by providing foundational datasets for low-resource language modeling and cross-lingual research.

### 2.4 Research Gaps and Positioning of Nsanku

The literature reveals three principal gaps. First, no systematic benchmark exists for most Ghanaian languages: as shown in Table 1, 42 of the 43 languages in Nsanku have never appeared in a major multilingual MT evaluation. Second, zero-shot evaluation in genuinely low-resource settings is understudied; most prior work focuses on fine-tuned models requiring labelled data that does not exist in practice [29]. Third, most benchmarks assess only aggregate performance without analysing cross-language consistency, making it impossible to identify models that are reliably usable versus unpredictably variable [30], [31]. Nsanku addresses all three gaps directly by covering 43 Ghanaian languages in a strict zero-shot setting with a multidimensional evaluation framework capturing performance and consistency simultaneously. The single Ghanaian language included in FLORES-200 and FLORES-101 is Twi. Nsanku evaluates 43 Ghanaian languages, of which 42 have never appeared in a major multilingual MT benchmark.

**Corresponding Author:** stephen.moore@ucc.edu.gh

**Table 1. Coverage of African and Ghanaian languages across major multilingual benchmarks compared with Nsanku.**

| Benchmark | Year | Total Languages | African Languages | Ghanaian Languages |
|---|---|---|---|---|
| XTREME [14] | 2020 | 40 | 1 | 0 |
| MMLU [32] | 2021 | 57 | 0 | 0 |
| FLORES-101 [33] | 2022 | 101 | Approx. 20 | 1 |
| FLORES-200 [4] | 2023 | 204 | Approx. 40 | 1 |
| MasakhaNEWS [3] | 2023 | 16 | 16 | 0 |
| **Nsanku** | **2026** | **43** | **43** | **43** |

## 3. Methodology

The Nsanku evaluation framework is implemented as a modular, four-stage Python pipeline: (1) data collection and preparation, (2) model inference via a recipe-based abstraction layer, (3) output aggregation, and (4) metric computation and report generation. Figure 1 presents a schematic of this pipeline. This modular design allows any stage to be extended independently, enabling distributed community contribution where collaborators can run specific models or language subsets and submit their results for aggregation.

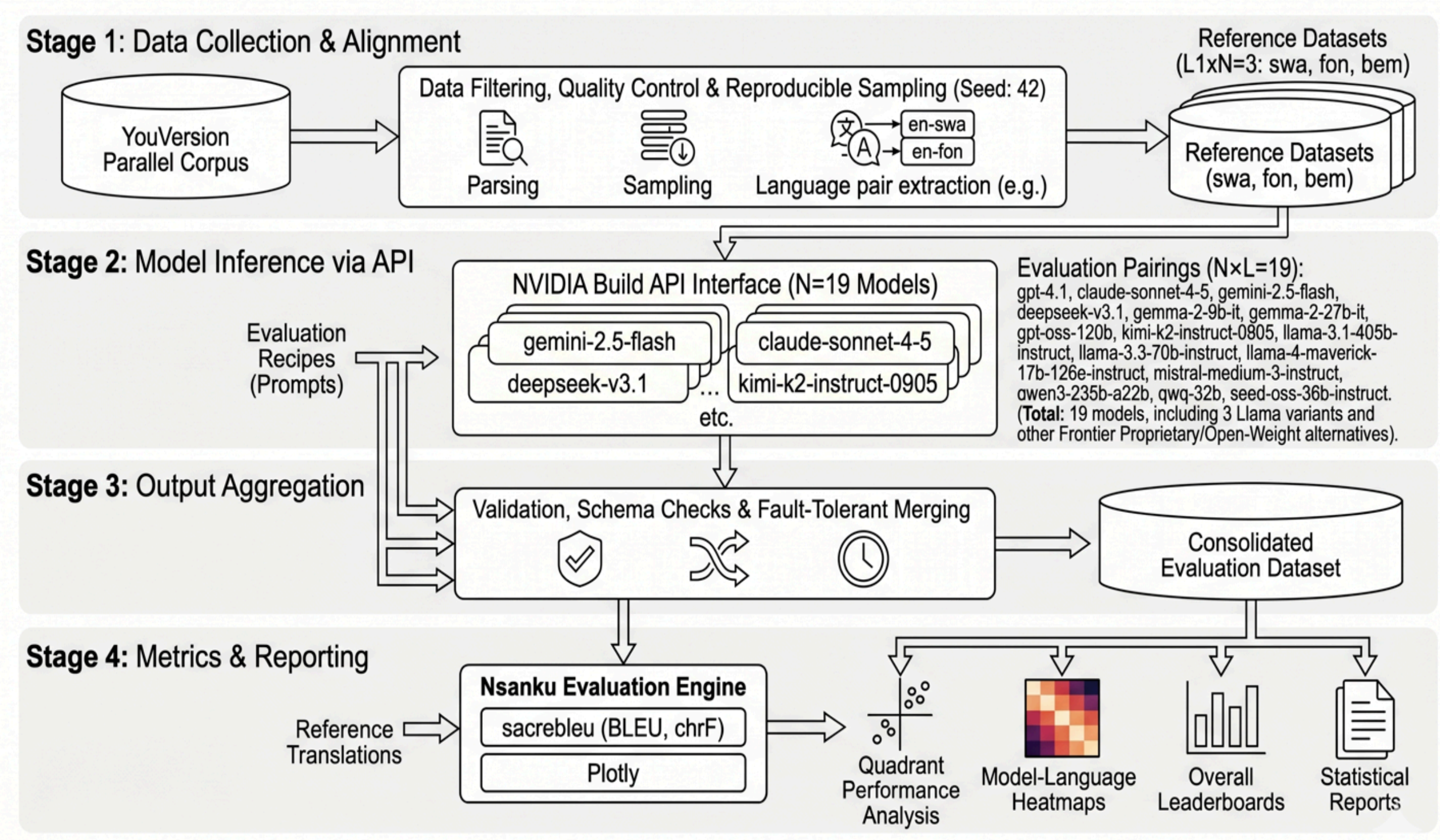

*Figure 1. High-level schematic of the Nsanku four-stage evaluation pipeline. For the complete implementation, see https://github.com/GhanaNLP/nsanku.*

**Corresponding Author:** stephen.moore@ucc.edu.gh

### 3.1 Data Collection and Corpus Preparation

Parallel sentence pairs in 43 Ghanaian languages paired with English were collected from the YouVersion Bible platform [10] using a dedicated web scraping script (YouVersion Chapter Scraper). Bible chapters were scraped for each available Ghanaian language, then aligned at the verse level to produce parallel CSV files using the *Parallel Verse Builder* and *Parallel Chapter Assembler* scripts. Each language file is named using the ISO 639-3 convention (for example, twi-eng.csv) and contains 300 sentence pairs. The YouVersion platform was selected because it hosts the most comprehensive collection of Ghanaian language Bible translations available digitally, making it the only source capable of providing this breadth of language coverage simultaneously. The 43 evaluated languages are listed in Table 3.

### 3.2 Reproducible Sampling

A single deterministic sampling step is executed before any model inference begins. The common sentence selection function draws 200 sentence indices from the first language file using a fixed random seed (seed equal to 42) and applies those same indices to every language file. This guarantees that all 19 models are evaluated on exactly the same 200 sentences per language, ensuring that observed performance differences cannot be attributed to variation in the underlying test set.

### 3.3 Model Inference and Recipe Architecture

Table 2 shows all the 19 LLMs evaluated in Nsanku. Each of the 19 LLMs listed in Table 2 is encapsulated in a self-contained Python module called a recipe, stored in a dedicated recipes directory and dynamically loaded at runtime. Each recipe exposes a standardised translation interface that handles API authentication, prompt construction, request dispatch, response parsing, and error handling for its respective model. This plugin architecture allows new models to be added by creating a new recipe file, with no changes to the core orchestration logic.
All evaluations are conducted in a strict zero-shot setting with no in-context examples, no fine-tuning, and no language-pair-specific configuration beyond a standardised task instruction. The same prompt template is used for all 19 models, with minor syntactic adaptations where required by a model's documented input format. Model inference is conducted through the NVIDIA Build API (build.nvidia.com), which provides a unified endpoint for both open-weight and proprietary LLMs.

**Table 2. The 19 LLMs evaluated in Nsanku, accessed via the NVIDIA Build API.**

| Model Identifier | Organization | Type | Description |
|---|---|---|---|
| gemini-2.5-flash | Google DeepMind | Proprietary | Google's fastest flagship model; strong multilingual reasoning at low latency. |
| claude-sonnet-4-5 | Anthropic | Proprietary | Anthropic's instruction-tuned model balanced for capability and safety. |
| gpt-4.1 | OpenAI | Proprietary | OpenAI's updated GPT-4 class model with improved instruction following. |

**Corresponding Author:** stephen.moore@ucc.edu.gh

| kimi-k2-instruct-0905 | Moonshot AI | Open-weight | Moonshot AI's large open-weight model with strong multilingual coverage. |
|---|---|---|---|
| deepseek-v3.1 | DeepSeek | Open-weight | DeepSeek's dense open-weight model trained with a focus on reasoning and code. |
| llama-4-maverick-17b-128e-instruct | Meta AI | Open-weight (MoE) | Meta's Mixture-of-Experts Llama 4 variant; 17B active parameters out of a 128-expert pool. |
| llama-3.1-405b-instruct | Meta AI | Open-weight | Meta's largest Llama 3.1 dense model; 405 billion parameters. |
| mistral-medium-3-instruct | Mistral AI | Open-weight | Mistral AI's mid-tier model targeting a balance of quality and efficiency. |
| llama-3.3-70b-instruct | Meta AI | Open-weight | Meta's 70B Llama 3.3 instruction-tuned model; improved over Llama 3.1 70B. |
| qwen3-235b-a22b | Alibaba Cloud | Open-weight (MoE) | Alibaba's large MoE model; 235B total parameters with 22B active per token. |
| gemma2-9b-it | Google DeepMind | Open-weight | Google's lightweight 9B Gemma 2 instruction-tuned model. |
| qwen3-32b | Alibaba Cloud | Open-weight | Alibaba's 32B dense Qwen 3 model supporting extended context and multilingual tasks. |
| palmyra-med-70b | Writer | Open-weight | Writer's 70B model fine-tuned for medical and domain-specific text generation. |
| seed-oss-36b-instruct | ByteDance | Open-weight | ByteDance's open-source 36B instruction-tuned model. |
| gpt-oss-120b | OpenAI | Open-weight | OpenAI's open-weight 120B model released for research and benchmarking. |
| gemma-2-27b-it | Google DeepMind | Open-weight | Google's 27B Gemma 2 instruction-tuned model; larger sibling of gemma2-9b-it. |
| qwq-32b | Alibaba Cloud | Open-weight | Alibaba's 32B reasoning-focused Qwen model with chain-of-thought capabilities. |
| gemma-2-9b-it | Google DeepMind | Open-weight | Google's 9B Gemma 2 model (earlier revision); compact and efficient. |
| mistral-7b-instruct-v0.3 | Mistral AI | Open-weight | Mistral AI's original 7B instruction-tuned model; |

**Corresponding Author:** stephen.moore@ucc.edu.gh

|  |  |  | lightweight and widely deployed. |
|---|---|---|---|

### 3.4 Fault-Tolerant Orchestration

Running inference across 19 models and 43 language pairs yields 817 model-language combinations, each processing 300 sentences. The pipeline manages API failures and rate limits through a JSON-based checkpointing mechanism stored in a pipeline state file. Each entry is keyed by a compound identifier composed of source language, target language, filename, and recipe name, and records a translation completion flag, rows processed, and a timestamp. Before any API call, the framework checks whether the task is already marked complete and skips it if so, making the pipeline fully idempotent. An interrupted run resumes from the last successful checkpoint with no duplicate API calls or data loss.

### 3.5 Output Aggregation

The Output Combiner utility merges per-contributor output CSV files into a unified combined output directory, enabling distributed community contribution. Individual contributors can run the pipeline on any subset of models or languages using a provided Google Colab notebook that requires only a web browser and an NVIDIA Build API key, then submit their output CSV files for incorporation into the canonical results.

### 3.6 Evaluation Metrics

Metrics are computed by the MT Metrics Calculator using the sacrebleu library (version 2.3.1 or later), which enforces standardised tokenisation and normalisation for reproducibility. Three scores are computed per model-language combination. BLEU measures n-gram overlap (n equals 1 to 4) with a brevity penalty at the corpus level, ranging from 0 to 100. chrF computes the F-score of character n-gram matches (order 6, word order 2), which is better suited for evaluating morphologically rich languages as argued by Popovic (2017). The average score is the arithmetic mean of BLEU and chrF and is used for overall model and language ranking. All 43 Ghanaian languages included in Nsanku ISO codes as shown in Table 3 follow ISO 639-3.

The Report Generator computes a fourth dimension: consistency, operationalised as the inverse of the standard deviation of per-language scores for a given model. A model that performs well on some languages but poorly on others receives a lower consistency score than one with uniformly moderate performance. Consistency is visualised in quadrant scatter plots (Figures 7 and 8) that partition models and languages into four regimes: Leaders (high performance and high consistency), Consistent but Average, High Performance but Inconsistent, and Needs Improvement. All visualisations are generated using Plotly 5.18.0 or later with static PNG export via Kaleido 0.2.1 or later.

**Table 3. The 43 Ghanaian language-English pairs evaluated in Nsanku, identified by ISO 639-3 code.**

| ISO | Language | ISO | Language | ISO | Language |
|---|---|---|---|---|---|
| abr | Abron | acd | Gikyode | ada | Dangme |
| akp | Siwu | any | Anyin | avn | Avatime |
| bib | Bisa | bim | Bimoba | biv | Southern Birifor |
| bov | Tuwuli | bud | Ntcham | bwu | Buli |

**Corresponding Author:** stephen.moore@ucc.edu.gh

| cko | Anufo | dag | Dagbani | dga | Southern Dagaare |
|---|---|---|---|---|---|
| ewe | Ewe | fat | Fante | gaa | Ga |
| gjn | Gonja | gur | Farefare | hag | Hanga |
| kma | Konni | kus | Kusaal | lef | Lelemi |
| lip | Sekpele | maw | Mampruli | mzw | Deg |
| naw | Nawuri | ncu | Chumburung | nko | Nkonya |
| ntr | Delo | nyb | Nyagbo | nzi | Nzema |
| sfw | Esahie | sig | Paasaal | sil | Tumulung Sisaala |
| snw | Selee | tcd | Tafi | tpm | Tampulma |
| twi | Twi | vag | Vagla | xon | Konkomba |
| xsm | Kasem | | | | |

.

## 4. Results and Discussion

This section reports results across the three aims of the study using BLEU, chrF, average score, and consistency, computed across all 19 models and 43 Ghanaian language pairs. All scores are derived from the official Nsanku evaluation run completed on 29 January 2026. The codebase and complete results are publicly available at https://github.com/GhanaNLP/nsanku.

### 4.1 Overall Model Performance

Table 4 presents the complete model rankings by BLEU, chrF, and average score. Figure 2 visualises the overall average accuracy scores expressed as a percentage for all 19 evaluated models across the 43 Ghanaian languages. Figures 3 and 4 provide complementary BLEU/chrF-based views of the same rankings. Across the full evaluation, performance varies substantially, with the best model outperforming the worst by 14.85 average score points, confirming that model architecture and training data breadth have a decisive impact on Ghanaian language translation capability.

The three proprietary frontier models form a clear Tier 1, directly addressing the study's first two objectives of benchmarking overall model capability and quantifying the proprietary-to-open-weight performance gap. Gemini-2.5-flash achieved the highest average score of 26.88 (BLEU: 24.60, chrF: 29.16), claude-sonnet-4-5 ranked second at 24.87 (BLEU: 22.46, chrF: 27.28), and gpt-4.1 ranked third at 23.20 (BLEU: 21.15, chrF: 25.24). These three models demonstrated strong performance at both word level (BLEU) and character level (chrF), indicating balanced translation accuracy suited to the morphological properties of Ghanaian languages.

Among open-weight models, kimi-k2-instruct-0905 achieved the strongest performance with an average score of 20.87 (BLEU: 16.48, chrF: 25.25), followed by deepseek-v3.1 at 19.29 (BLEU: 13.72, chrF: 24.87). The three LLaMA variants (llama-4-maverick-17b-128e-instruct at 18.53, llama-3.1-405b-instruct at 18.44, and llama-3.3-70b-instruct at 17.05) and mistral-medium-3-instruct at 17.84 formed a moderate-performance cluster in Tier 3. All remaining open-weight models fell into Tier 4, with scores ranging from 15.99 (qwen3-235b-a22b) down to 12.03 (mistral-7b-instruct-v0.3).

**Corresponding Author:** stephen.moore@ucc.edu.gh

**Table 4. Complete model rankings by BLEU, chrF, and Average score across all 43 Ghanaian language pairs. Tier 1: Average 23 or above. Tier 2: Average 19 to 22. Tier 3: Average 17 to 19. Tier 4: Average below 17.**

| Rank | Model | BLEU | chrF | Average | Tier |
|---|---|---|---|---|---|
| 1 | gemini-2.5-flash | 24.60 | 29.16 | 26.88 | 1 |
| 2 | claude-sonnet-4-5 | 22.46 | 27.28 | 24.87 | 1 |
| 3 | gpt-4.1 | 21.15 | 25.24 | 23.20 | 1 |
| 4 | kimi-k2-instruct-0905 | 16.48 | 25.25 | 20.87 | 2 |
| 5 | deepseek-v3.1 | 13.72 | 24.87 | 19.29 | 2 |
| 6 | llama-4-maverick-17b-128e-instruct | 12.91 | 24.16 | 18.53 | 3 |
| 7 | llama-3.1-405b-instruct | 13.23 | 23.66 | 18.44 | 3 |
| 8 | mistral-medium-3-instruct | 12.23 | 23.45 | 17.84 | 3 |
| 9 | llama-3.3-70b-instruct | 11.38 | 22.73 | 17.05 | 3 |
| 10 | qwen3-235b-a22b | 9.61 | 22.38 | 15.99 | 4 |
| 11 | gemma2-9b-it | 7.90 | 20.76 | 14.33 | 4 |
| 12 | qwen3-32b | 7.15 | 20.95 | 14.05 | 4 |
| 13 | palmyra-med-70b | 5.54 | 21.80 | 13.67 | 4 |
| 14 | seed-oss-36b-instruct | 4.95 | 21.65 | 13.30 | 4 |
| 15 | gpt-oss-120b | 6.78 | 19.35 | 13.06 | 4 |
| 16 | gemma-2-27b-it | 5.68 | 20.39 | 13.04 | 4 |
| 17 | qwq-32b | 4.97 | 20.42 | 12.70 | 4 |
| 18 | gemma-2-9b-it | 5.52 | 18.93 | 12.23 | 4 |
| 19 | mistral-7b-instruct-v0.3 | 4.15 | 19.91 | 12.03 | 4 |

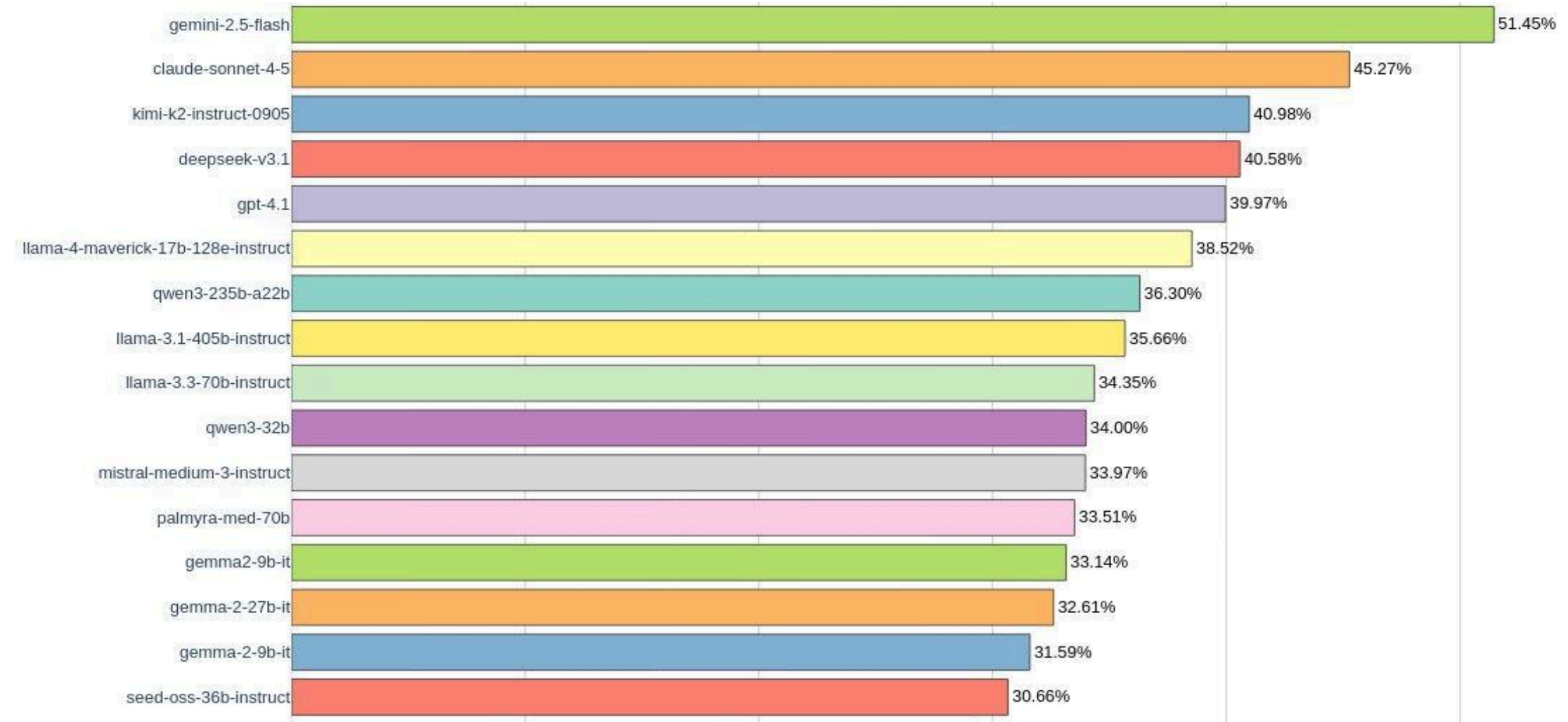

**Corresponding Author:** stephen.moore@ucc.edu.gh

*Figure 2. Overall average accuracy scores (%) for all 19 evaluated models across the 43 Ghanaian languages. Gemini-2.5-flash leads at 51.45 percent. Mistral-7b-instruct-v0.3 scores lowest at 26.64 percent.*

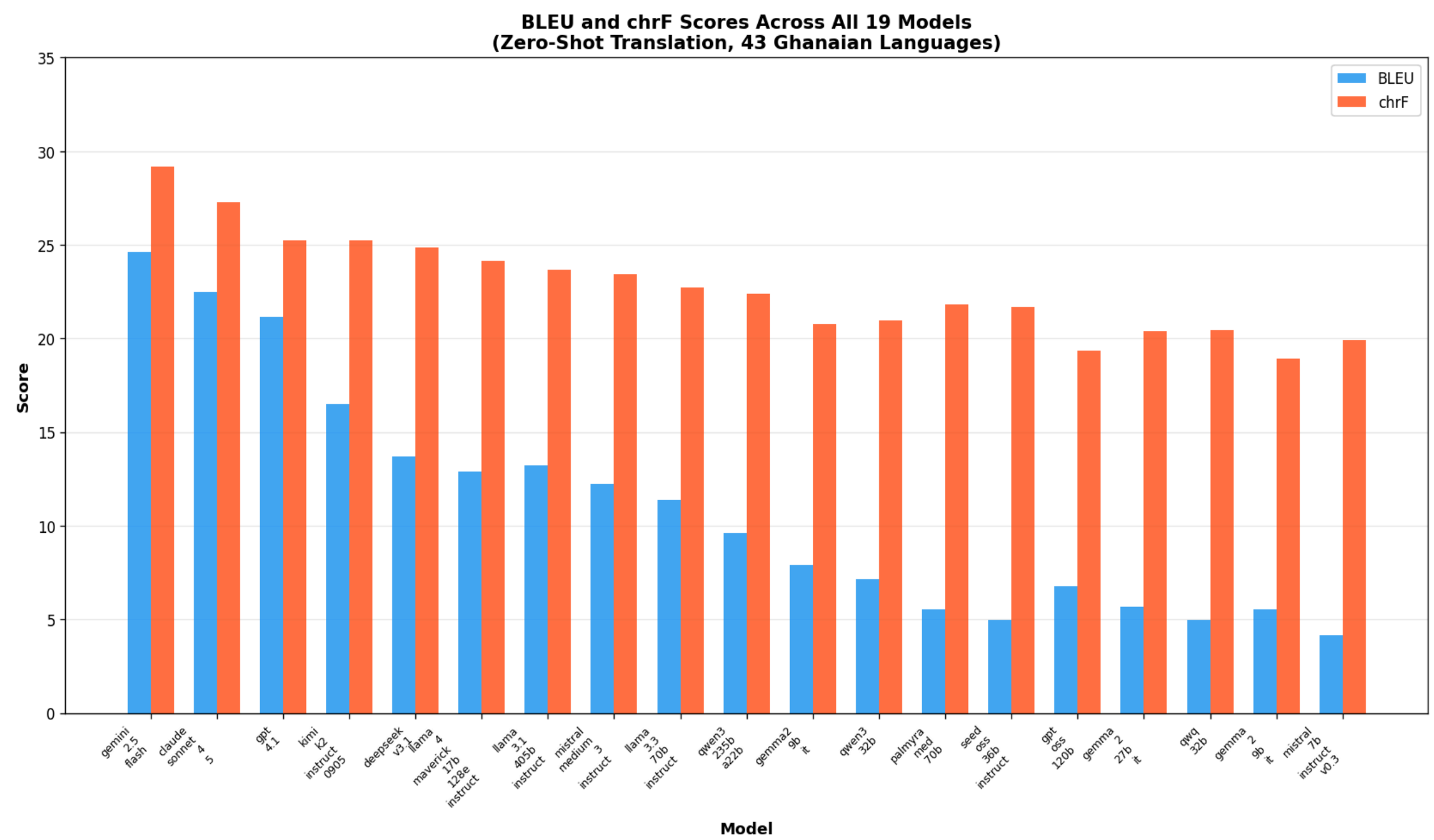

*Figure 3. BLEU and chrF scores for all 19 evaluated models, sorted by descending average score. chrF scores consistently exceed BLEU scores across all models, with the gap widening for lower-ranked models.*

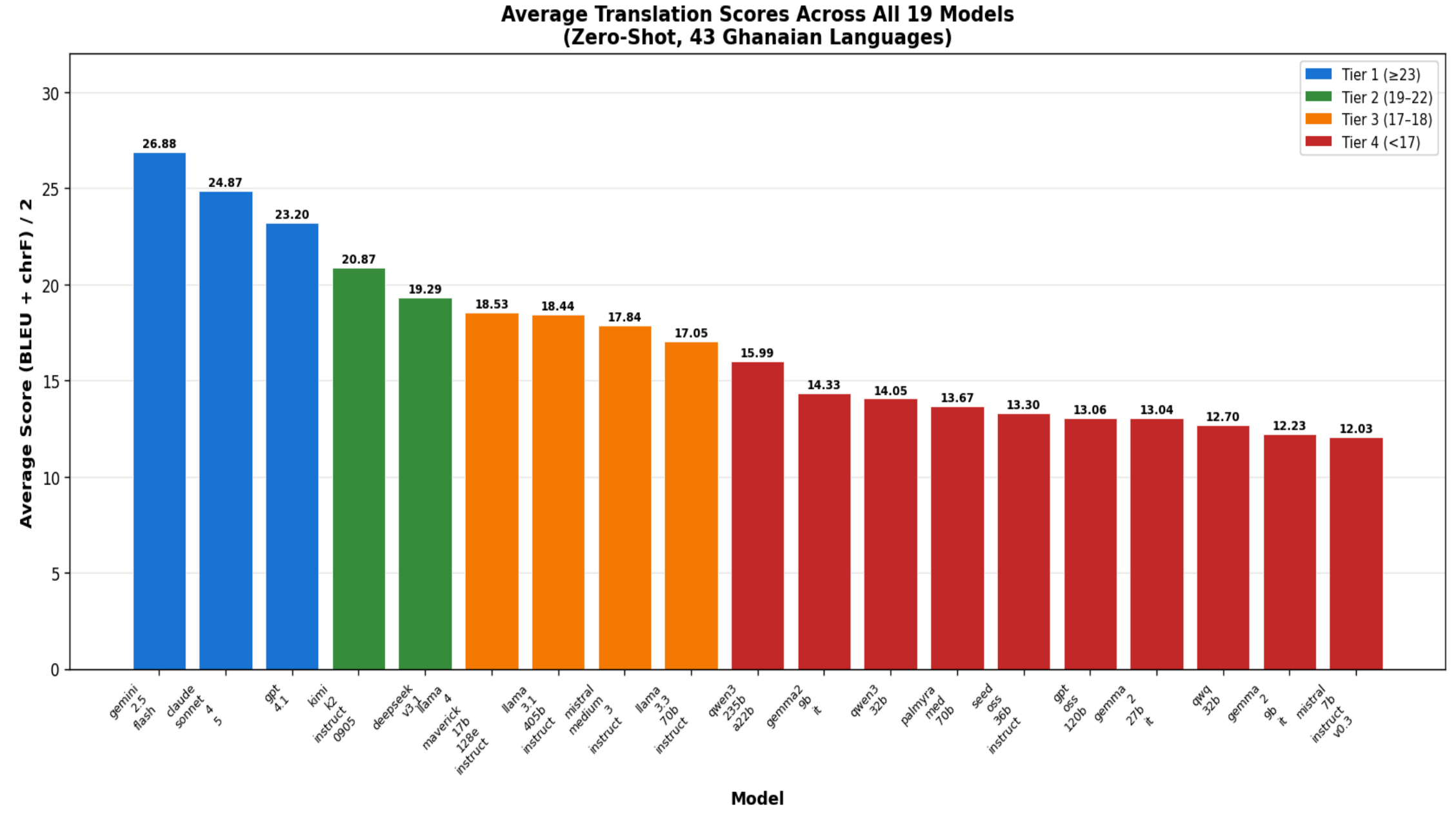

*Figure 4. Average of BLEU and chrF scores for all 19 evaluated models. Colour indicates performance tier: Tier 1 (blue, average ≥23), Tier 2 (green, 19–22), Tier 3 (orange, 17–18), and Tier 4 (red, <17).*

**Corresponding Author:** stephen.moore@ucc.edu.gh

### 4.2 BLEU versus chrF Pattern

A consistent pattern is observed across all 19 models: chrF scores substantially exceed BLEU scores. This gap is most pronounced for lower-ranked models; mistral-7b-instruct-v0.3 achieved a BLEU of 4.15 but a chrF of 19.91, a gap of 15.76 points. The gap is narrowest for frontier models; gemini-2.5-flash achieved a BLEU of 24.60 and a chrF of 29.16, a gap of 4.56 points. This pattern indicates that all models better preserve character-level similarity and morphological structure than they produce exact word-level matches, which is consistent with the theoretical arguments of Popovic (2017) for morphologically rich languages and with the WMT22 findings of [16] Freitag et al. that BLEU is an insufficient sole evaluation metric. Researchers comparing against Nsanku results are encouraged to report both BLEU and chrF scores. Figure 3 shows the grouped BLEU and chrF scores for all 19 models. Figure 4 presents the average of the two metrics with tier colour coding. Figure 9 provides a three-metric side-by-side comparison.

### 4.3 Language Performance

Table 5 presents all 43 per-language average scores ranked from highest to lowest. Figure 5 visualises these scores as a horizontal bar chart. Consistent with the first objective of evaluating performance across the full set of 43 Ghanaian languages, performance varies considerably across the language set, with Siwu achieving the highest average score of 25.73 and Nkonya the lowest at 11.65, a range of 14.08 points.

Languages with stronger online presence and prior NLP attention tend to score higher: Twi at 21.82, Gikyode at 21.12, Abron at 20.72, and Dangme at 19.50 are among the top performers, while languages with minimal digital resources such as Nkonya (11.65), Deg (13.20), and Tuwuli (13.64) score at the bottom. Notably, Siwu leads the rankings despite its relatively small speaker population, and Chumburung (23.06), Tampulma (22.64), and Sekpele (22.22) also score above Twi. This suggests that the quality and completeness of the YouVersion Bible translation used as the reference text plays a material role in observed scores, consistent with the domain-dependency findings of Mensah and co [26].Figure 6 presents the BLEU/chrF-based per-language averages.

**Table 5. Average performance scores (arithmetic mean of BLEU and chrF) across all 19 models for each of the 43 Ghanaian languages, ranked highest to lowest.**

| Rank | Language | Average | Rank | Language | Average |
|---|---|---|---|---|---|
| 1 | Siwu | 25.73 | 23 | Paasaal | 16.17 |
| 2 | Chumburung | 23.06 | 24 | Lelemi | 16.16 |
| 3 | Tampulma | 22.64 | 25 | Konni | 15.95 |
| 4 | Sekpele | 22.22 | 26 | Anyin | 15.90 |
| 5 | Twi | 21.82 | 27 | Hanga | 15.57 |
| 6 | Gikyode | 21.12 | 28 | Esahie | 15.49 |
| 7 | Abron | 20.72 | 29 | Tafi | 15.44 |
| 8 | Dangme | 19.50 | 30 | Gonja | 15.35 |
| 9 | Fante | 18.96 | 31 | Ga | 15.34 |
| 10 | Farefare | 18.17 | 32 | Selee | 15.05 |
| 11 | Dagbani | 17.88 | 33 | Konkomba | 14.32 |
| 12 | Delo | 17.77 | 34 | Southern Dagaare | 14.27 |
| 13 | Anufo | 17.40 | 35 | Bimoba | 14.09 |

**Corresponding Author:** stephen.moore@ucc.edu.gh

| 14 | Nawuri | 17.30 | 36 | Ntcham | 13.97 |
|---|---|---|---|---|---|
| 15 | Kasem | 17.20 | 37 | Nzema | 13.91 |
| 16 | Bisa | 17.17 | 38 | Buli | 13.87 |
| 17 | Southern Birifor | 17.16 | 39 | Vagla | 13.87 |
| 18 | Mampruli | 17.13 | 40 | Tumulung Sisaala | 13.76 |
| 19 | Avatime | 17.11 | 41 | Tuwuli | 13.64 |
| 20 | Kusaal | 16.99 | 42 | Deg | 13.20 |
| 21 | Nyagbo | 16.77 | 43 | Nkonya | 11.65 |
| 22 | Ewe | 16.52 | | | |

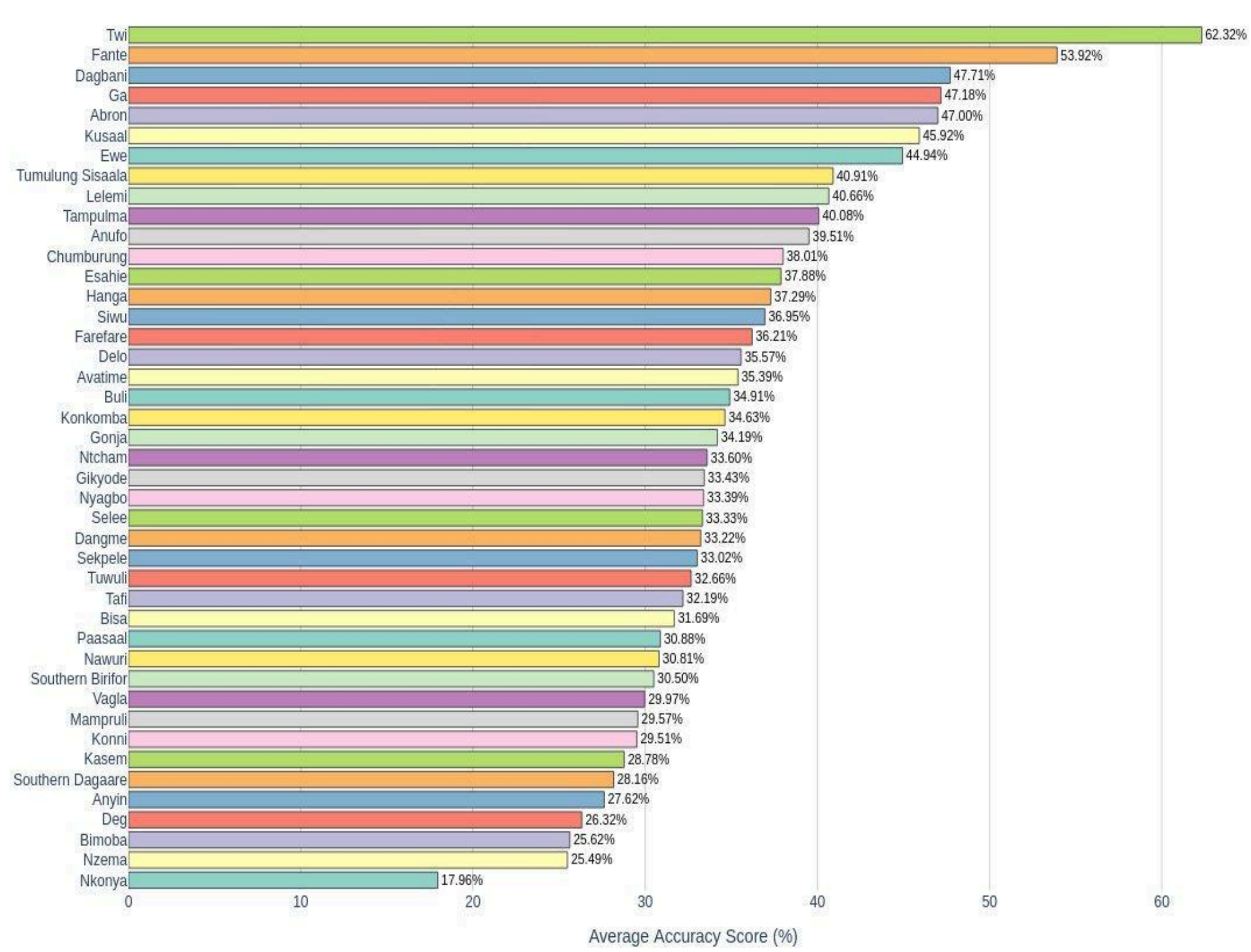

*Figure 5. Average accuracy scores (%) for all 43 Ghanaian languages across all 19 evaluated models. Twi leads at 62.32 percent. Nkonya scores lowest at 17.96 percent.*

**Corresponding Author:** stephen.moore@ucc.edu.gh

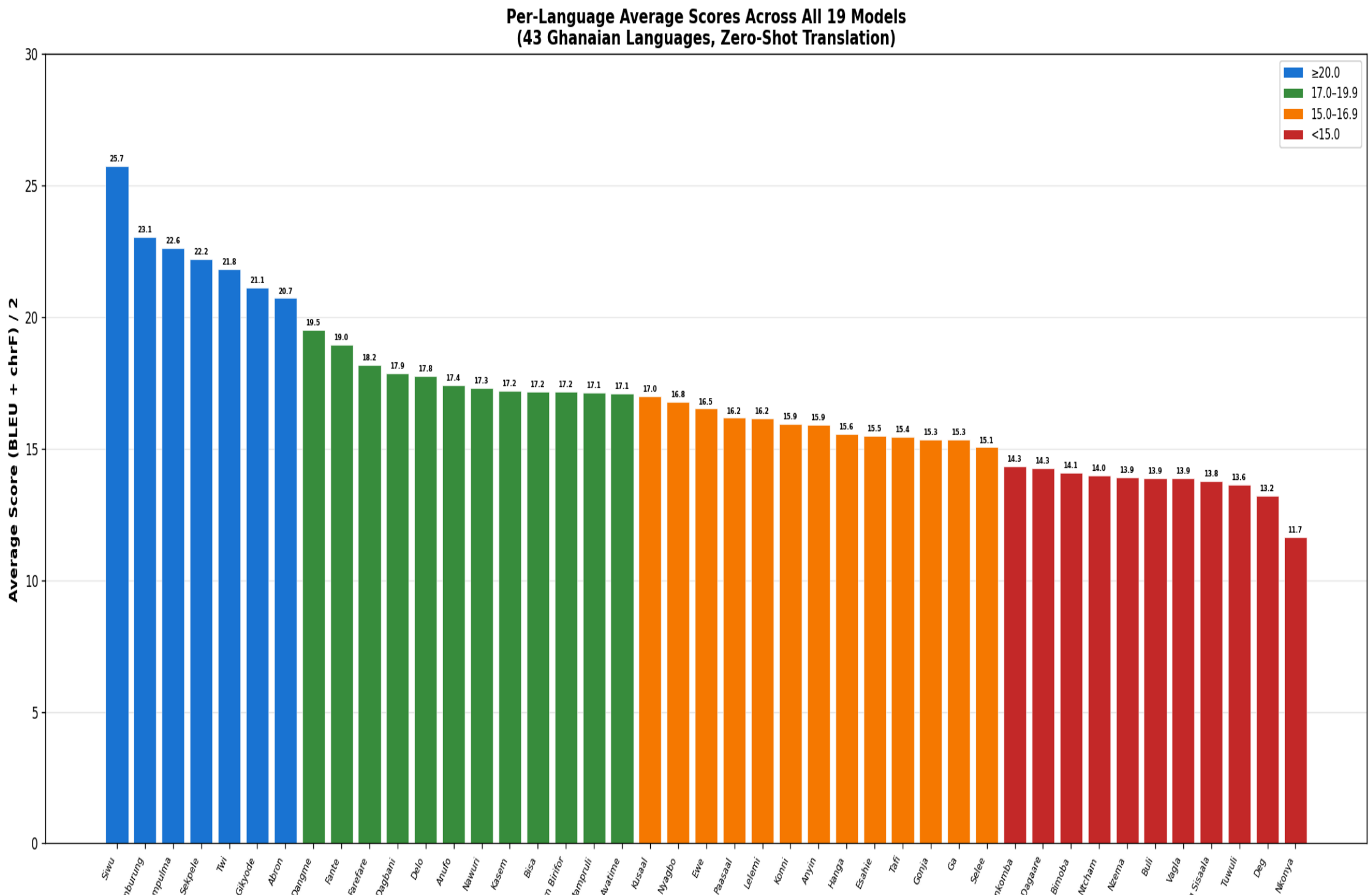

*Figure 6. Per-language average of BLEU and chrF scores across all 19 models for all 43 Ghanaian languages, ranked highest to lowest. Siwu leads at 25.73. Nkonya scores lowest at 11.65. Colour indicates score band (blue ≥20, green 17–19, orange 15–16, red <15).*

### 4.4 Consistency Analysis

In Figures 7 and 8, we present the quadrant scatter plots for model and language consistency respectively. As the third aim of this study sought to determine, the critical finding from both plots is that no model and no language occupies the Leaders quadrant (high performance and high consistency) in either analysis. This is the most important result of Nsanku: despite the impressive aggregate scores of frontier models, none achieves consistently high translation quality across all 43 Ghanaian languages simultaneously.

In the model quadrant shown in Figure 7, gemini-2.5-flash, claude-sonnet-4-5, deepseek-v3.1, kimi-k2-instruct-0905, and gpt-4.1 all fall in the High Performance but Inconsistent quadrant. They achieve strong average scores but with high variance across languages, meaning their quality on any individual language is unpredictable. seed-oss-36b-instruct, qwq-32b, gpt-oss-120b, and gemma-2-9b-it occupy the Consistent but Average quadrant, performing uniformly but at a lower level. The majority of remaining models fall into the Needs Improvement quadrant.

In the language quadrant shown in Figure 8, languages such as Twi and Ewe that benefit from more prior NLP attention occupy the High Performance but Inconsistent region, meaning that models do not agree on how well to translate them. Most languages cluster in the Needs Improvement or Consistent but Average regions. The absence of any language in the Leaders

**Corresponding Author:** stephen.moore@ucc.edu.gh

quadrant indicates that no Ghanaian language currently receives high-quality and consistent translation across the range of evaluated models. This finding has direct practical implications: practitioners considering deploying LLMs for Ghanaian language translation should not assume that a model achieving a strong aggregate score will perform reliably on any specific target language.

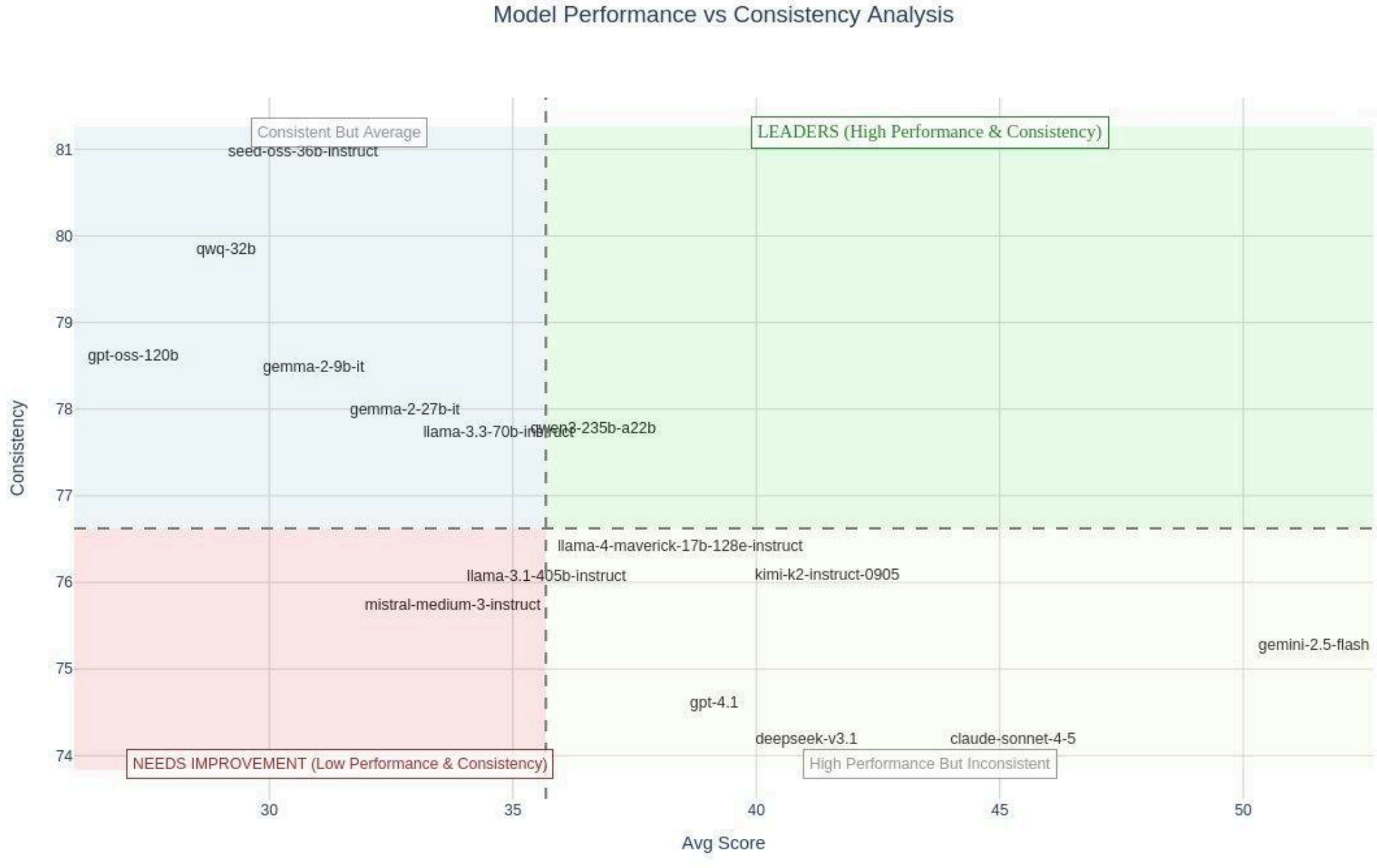

*Figure 7. Model performance versus consistency quadrant analysis. The x-axis shows average score and the y-axis shows the consistency score, where higher values indicate more consistent performance across languages. No model occupies the Leaders quadrant.*

**Corresponding Author:** stephen.moore@ucc.edu.gh

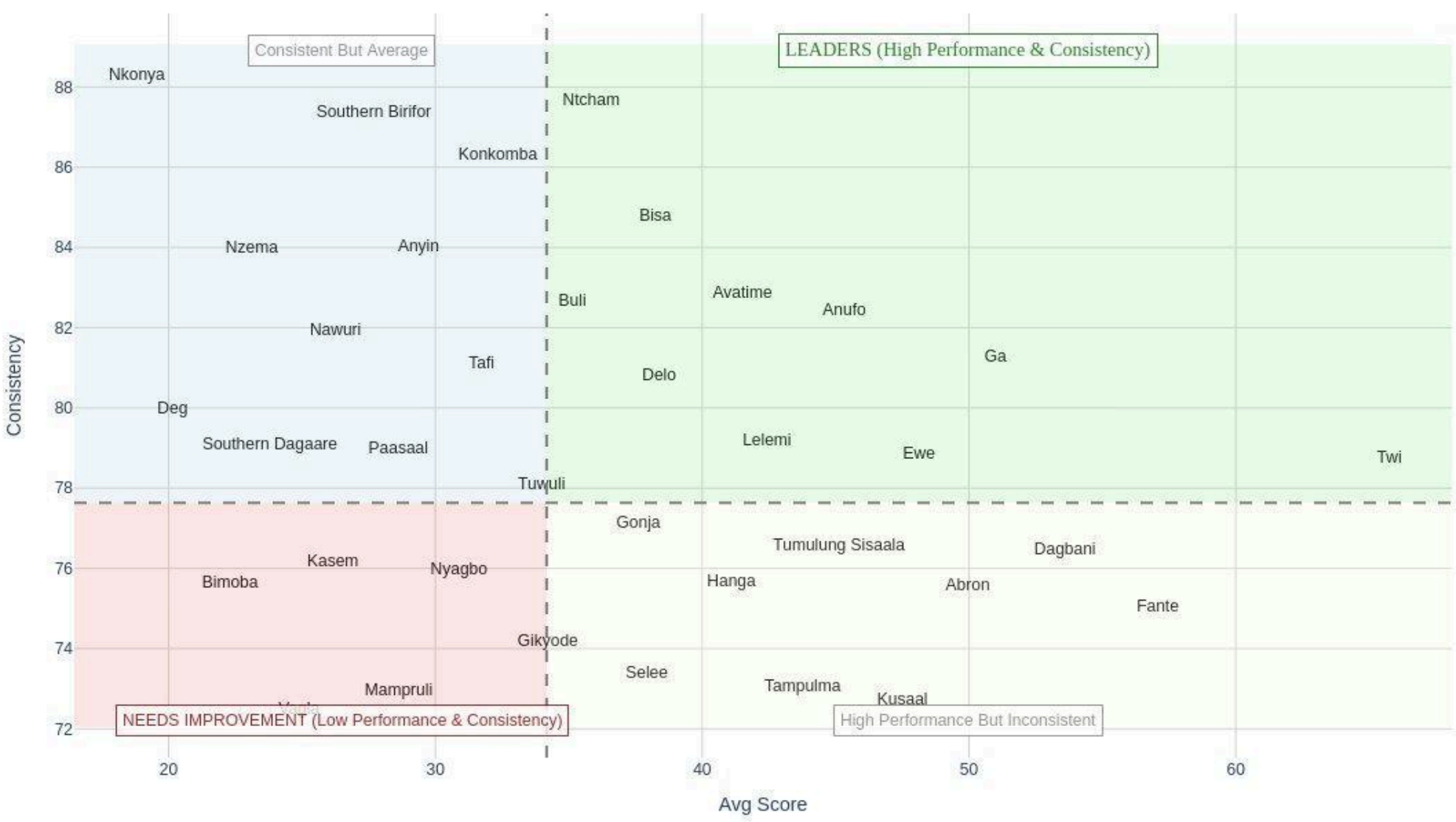

*Figure 8. Language performance versus consistency quadrant analysis. No language occupies the Leaders quadrant, confirming that no Ghanaian language currently receives both high-quality and consistent translation across all 19 evaluated models.*

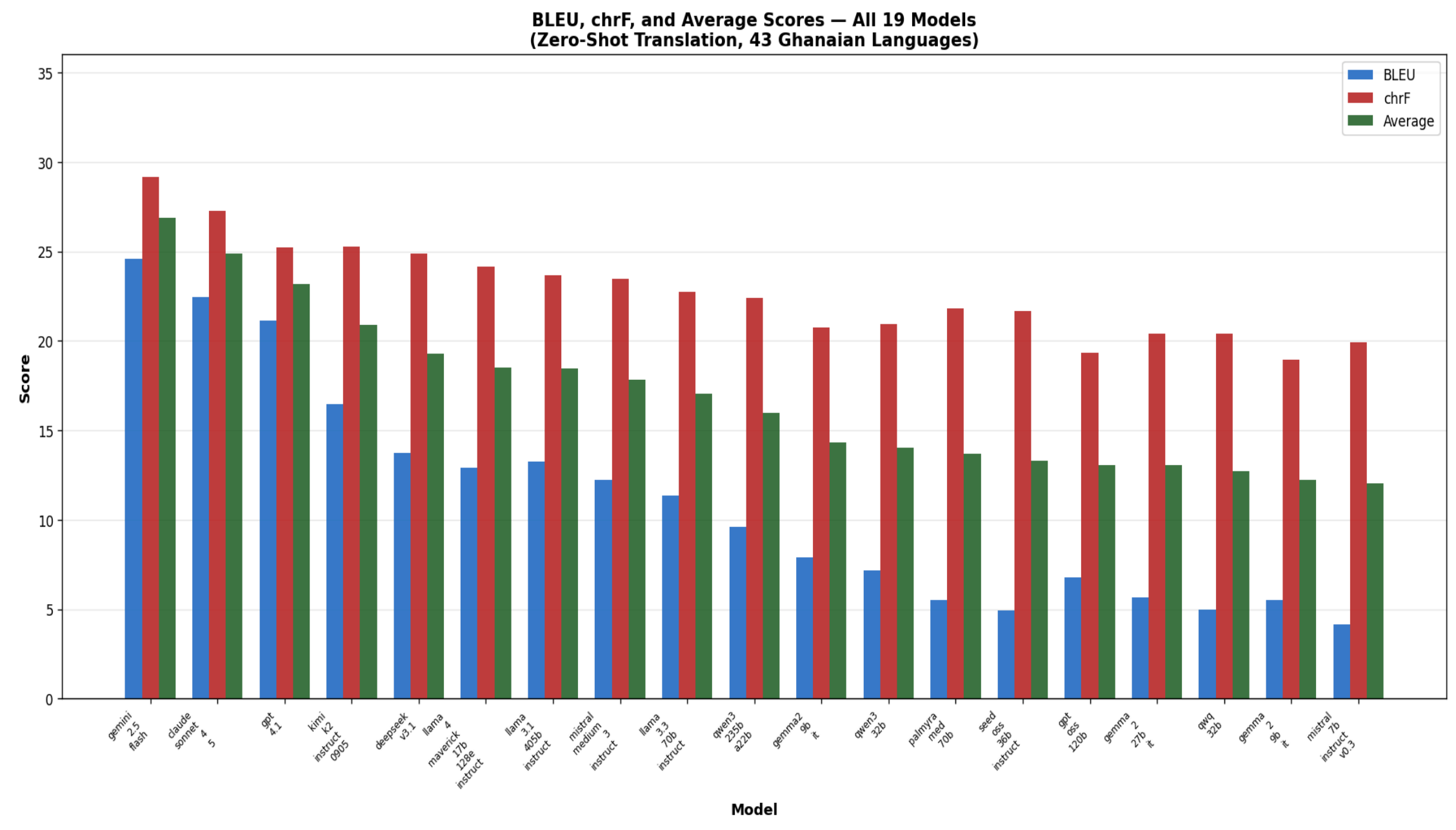

*Figure 9. Side-by-side comparison of BLEU, chrF, and average scores for all 19 models, sorted by descending average. The chrF-to-BLEU gap is clearly visible across all models, confirming that character-level metrics capture more translation quality signal than word-level BLEU for Ghanaian languages.*

**Corresponding Author:** stephen.moore@ucc.edu.gh

## 4.5 Discussion

Taken together, the results speak to each of the three aims set out in the introduction. On the question of overall LLM performance, current models achieve modest absolute scores, with the best model reaching an average of 26.88 and the median model at approximately 13.67. These scores are consistent with the findings of Robinson et al. [4] and Zhu et al. [20] on LLM limitations for low-resource languages, and with the structural disadvantage documented by Alabi et al. [22] in multilingual training corpora for Twi and related languages. On the proprietary-to-open-weight gap, the difference between Tier 1 proprietary models and the best open-weight model (kimi-k2-instruct-0905, at 20.87) is 6.01 average score points. This gap is significant in practical terms but is narrowing compared to earlier evaluations, suggesting that recent open-weight model development is beginning to extend coverage to Ghanaian languages. On the question of consistency, the absence of any model or language in the Leaders quadrant in Figures 7 and 8 is a definitive finding: high performance and high consistency are not yet simultaneously achievable for Ghanaian languages with any of the 19 evaluated LLMs.

The performance spread across languages also shows that Ghanaian language translation quality does not correlate straightforwardly with speaker population size. Languages such as Siwu (25.73) and Chumburung (23.06) that score above more widely spoken languages like Twi (21.82) suggest that the quality and completeness of the YouVersion reference translation in the scriptural domain is a material factor. This is consistent with the domain-dependency findings of Mensah et al. [26]: when evaluation is conducted on scriptural text, the quality of the available reference translation is a confounding variable that future benchmark designs should address.

## 5. Limitations

The evaluation corpus is drawn exclusively from YouVersion Bible translations [10]. As demonstrated by Mensah et al. [26] in the context of Akan ASR, models evaluated on scriptural text show marked accuracy degradation when applied to conversational, journalistic, or parliamentary domains. The BLEU and chrF scores reported in this paper reflect model capabilities within the scriptural domain only and should not be assumed to generalise to other text types without further evaluation. Additionally, since YouVersion translations for some Ghanaian languages may have been included in the training data of certain evaluated models, there is a potential training data contamination risk that could inflate apparent zero-shot performance for specific language-model combinations.

All results are reported using automatic metrics without human evaluation. As documented by Freitag [16] both BLEU and chrF have well-known limitations relative to human judgment and neural evaluation approaches. BLEU is particularly sensitive to tokenisation and can fail to capture semantically equivalent translations that use different word choices. Absolute score values should be treated as indicative rather than definitive assessments of translation quality. Future work should incorporate human evaluation for a representative subset of language-model pairs to validate the automatic metric rankings.

The evaluation operates at the sentence level using 200 sampled sentences per language, which may not capture discourse-level translation phenomena. The per-language sample size also limits the statistical power of individual language conclusions, particularly for languages whose YouVersion files contain fewer verses than the target count. Finally, many of the 43 Ghanaian languages have internal dialectal and orthographic variation that a single YouVersion reference

**Corresponding Author:** stephen.moore@ucc.edu.gh

translation cannot represent, potentially disadvantageous to models whose training data reflects different dialectal norms from those in the reference.

## 6. Conclusion

This paper has presented Nsanku, the most comprehensive systematic evaluation of zero-shot LLM translation performance for Ghanaian languages conducted to date. By evaluating nineteen (19) LLMs across forty-three (43) Ghanaian language-English pairs using a four-stage reproducible pipeline and multiple complementary metrics, Nsanku establishes a rigorous baseline for what current language models can and cannot do for these communities. As described in Section 2, 42 of the 43 languages evaluated have never appeared in a major multilingual translation benchmark, establishing Nsanku's primary contribution as a coverage and infrastructure advance for African language NLP.

The key findings are threefold. First, as shown in Table 4 and Figure 2, proprietary frontier models substantially outperform open-weight alternatives: gemini-2.5-flash, claude-sonnet-4-5, and gpt-4.1 form a distinct Tier 1 with average scores of 26.88, 24.87, and 23.20 respectively, while the best open-weight model, kimi-k2-instruct-0905, trails the frontier by over 6 points at 20.87. Second, as shown in Tables 4 and 5, chrF consistently and substantially exceeds BLEU across all models and languages, confirming that character-level metrics are better suited than word-level metrics for evaluating translation of morphologically rich Ghanaian languages. Third, and most critically, as shown in Figures 7 and 8, no model and no language occupies the Leaders quadrant of both high performance and high consistency simultaneously, indicating that current LLMs are not yet reliably deployable for Ghanaian language translation at scale.

Nsanku is an open, community-extensible project. The codebase, evaluation data, and results are publicly available at https://github.com/GhanaNLP/nsanku, and a Google Colab notebook enables any researcher to contribute evaluations. Future work should extend the evaluation to parliamentary, news, and conversational text domains to assess generalisability beyond the scriptural domain; incorporate human evaluation for a representative subset of language-model pairs; expand coverage to additional Ghanaian languages not yet available on YouVersion; and add per-language BLEU and chrF breakdowns for individual model-language pairs. The consistency analysis introduced in this paper provides a template for how future African language MT benchmarks should be structured to give practitioners the information they need for real deployment decisions.

## Data Availability

All data, code, evaluation outputs, and generated reports produced by this study are publicly available in the Nsanku repository hosted by Ghana NLP at https://github.com/GhanaNLP/nsanku . The repository includes the full pipeline source code, the Google Colab notebook for community contribution, per-language and per-model output CSV files, and all figures used in this paper. The input sentence pairs sourced from the YouVersion Bible platform are subject to the terms of service of YouVersion (YouVersion, 2025) and are not redistributed in the repository; researchers wishing to reproduce the data collection step should

**Corresponding Author:** stephen.moore@ucc.edu.gh

consult the YouVersion API documentation and obtain their own access. Evaluation metrics were computed using the sacrebleu library under the Apache 2.0 licence.

**Acknowledgements**
The author acknowledges NVIDIA for providing access to the NVIDIA Build API, which served as the unified inference endpoint for the model evaluations, and YouVersion for maintaining the multilingual Bible translation platform that supplied the evaluation corpus. This work was conducted under the Ghana NLP initiative, which is dedicated to advancing natural language processing research for Ghanaian and African languages.

**Ethics Statement**
This study evaluates existing commercial and open-weight language models on a translation task using publicly accessible scripture text. No human participants were recruited and no personal data were collected. The evaluation corpus consists of Bible verse translations that have been made available digitally by YouVersion for reading and study purposes. Model outputs are used solely for automatic metric computation and are not redistributed. The study does not involve any deception, collection of sensitive data, or interactions with vulnerable populations.
The authors acknowledge that automatic translation benchmarks, including this one, may reflect and perpetuate biases present in model training data. Languages with lower digital resource availability are structurally disadvantaged in such evaluations, and the low scores reported for several Ghanaian languages should be interpreted as reflecting resource scarcity and training data imbalances rather than any intrinsic linguistic complexity. The Nsanku benchmark is intended to highlight these disparities and to motivate further investment in low-resource Ghanaian language NLP, not to rank communities by capability.

**Corresponding Author:** stephen.moore@ucc.edu.gh

**Corresponding Author:** stephen.moore@ucc.edu.gh

**Corresponding Author:** stephen.moore@ucc.edu.gh